\documentclass{article}
\usepackage{spconf,amsmath,graphicx,mathrsfs,booktabs,amssymb}
\usepackage{pifont}
\usepackage{multirow,makecell}
\usepackage{caption} 
\usepackage{cite}
\usepackage{color,soul}
\usepackage{changes}

\title{Video Anomaly Detection via prediction network with enhanced spatio-temporal memory exchange}
\name{Guodong Shen, Yuqi Ouyang, and Victor Sanchez}
\address{Department of Computer Science, University of Warwick, Coventry, UK}

\begin{document}
%
\maketitle
\begin{abstract}
Video anomaly detection is a challenging task because most anomalies are scarce and non-deterministic. Many approaches investigate the reconstruction difference between normal and abnormal patterns, but neglect that anomalies do not necessarily correspond to large reconstruction errors. To address this issue, we design a   Convolutional LSTM Auto-Encoder prediction framework with enhanced spatio-temporal memory exchange   using bi-directionalilty and a higher-order mechanism. The bi-directional structure promotes learning the temporal regularity through forward and backward predictions. The unique higher-order mechanism further strengthens spatial information interaction between the encoder and the decoder. Considering the limited receptive fields in Convolutional LSTMs, we also introduce an attention module to highlight informative features for prediction. Anomalies are eventually identified by comparing the frames with their corresponding predictions. Evaluations on three popular benchmarks show that our framework  outperforms most existing prediction-based anomaly detection methods.
\end{abstract}
\begin{keywords}
video anomaly detection, prediction, ConvLSTM Auto-Encoder, bi-directional, attention
\end{keywords}
\section{Introduction}
\label{sec:intro}

Video anomaly detection aims to detect abnormal patterns in motion and appearance. The main challenges of this task come from the rarity and diversity of anomalies in the real world. 
To this end, recent works adopt unsupervised learning algorithms, i.e., modeling normal patterns from a training set and then discriminating anomalies based on the discrepancy from these normal patterns. One  representative strategy is measuring the reconstruction error produced by Auto-Encoders (AEs) under the assumption that anomalies would lead to outliers in the image or sparse coding space \cite{zaheer2020old, ouyang2021video}. However, despite its prevalence, this strategy exposes two significant limitations. First, the assumption may not necessarily hold  due to the high capacity of neural networks to reconstruct anomalies precisely. Second, the temporal dynamics of videos, which help to  uncover motion irregularities, are usually confined to alternative representations, e.g., optical flow.

Prediction-based approaches can overcome the aforementioned limitations by considering anomalies as unexpected events. Specifically, during training, they enforce a model  to generate accurate predictions (i.e., future frames) for normal events.  Once trained, they infer anomalies from the dissimilarity between prediction and ground truth. This process exploits both underlying spatial and temporal dynamics from historical observations so that anomalies result in poor predictions.

\begin{figure}[t]

\begin{minipage}[b]{1.0\linewidth}
  \centering
  \centerline{\includegraphics[width=8.5cm]{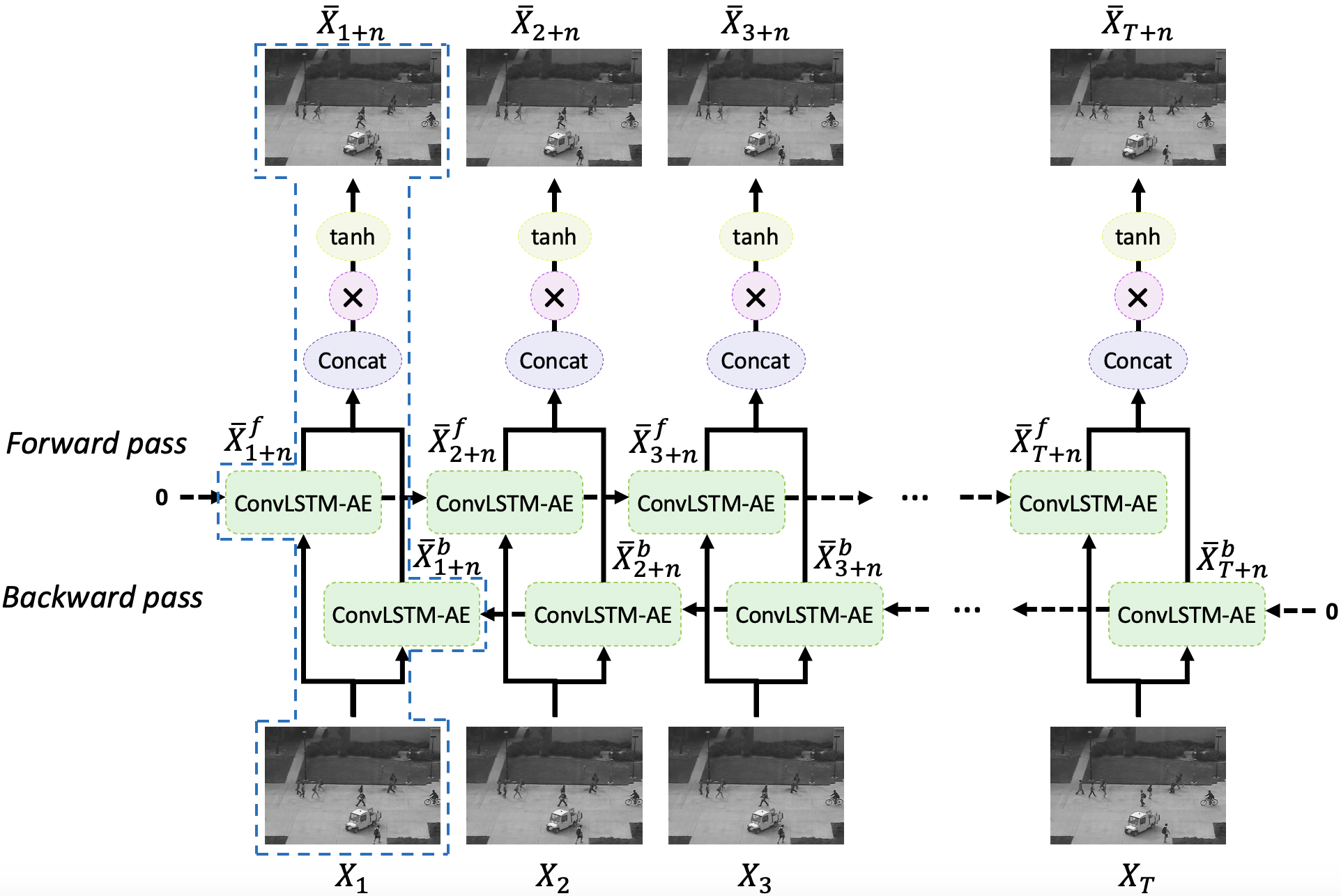}}
\end{minipage}

\caption{\small The pipeline of the proposed framework (unfolded over time). $\otimes$ denotes the  convolution operation. The dashed line highlights its fundamental structure.}
\label{fig:architecture}
\vspace{-3mm}
\end{figure}

In general, prediction-based approaches  can be categorized into two classes. 1) U-Net \cite{ronneberger2015u} based methods, which commonly stack consecutive frames to form spatio-temporal cubes that serve as input to U-Nets to predict subsequent frames \cite{liu2018future,lai2020video,chen2020anomaly}. Naively stacking frames, however, can  weaken the correlation between adjacent frames and burden the encoders in U-Nets. 2) Convolutional Long Short-term Memory (ConvLSTM) \cite{xingjian2015convolutional} based methods. Inspired by natural language processing (NLP), these methods employ recurrent frameworks to sequentially process data to generate future frames  \cite{luo2017remembering,wang2018abnormal,lee2019bman,song2019learning}.  However, most of these methods arrange and stack ConvLSTM layers like traditional LSTM layers, thus neglecting their difference in receptive fields. Moreover, both classes of methods usually require either adversarial learning \cite{liu2018future,song2019learning} or extra reconstruction tasks  \cite{lai2020video,luo2017remembering} to attain strong performance.

Within the context of ConvLSTM-based methods, this paper further bridges the gap between NLP and  prediction-based video anomaly detection by presenting a spatio-temporal memory-enhanced ConvLSTM-AE framework using bi-directionalilty  and a higher-order  mechanism \cite{yu2017long}. Compared to the uni-directionality adopted by previous works,  the bi-directionality supports the reverse processing  of the data. Our framework is thus more responsive to the temporal dimension of videos, which typically reflects motion irregularities.  Motivated by the higher-order  mechanism and PredRNN in \cite{wang2017predrnn}, we  devise a novel spatial higher-order ConvLSTM to boost spatial information exchange between the encoder and the decoder. Furthermore, to cope with the inflexible receptive fields in ConvLSTMs,  we also introduce  an effective attention module to dynamically highlight the features that are more informative for predicting future frames. Compared to the state-of-the-art (SOTA), the proposed framework demonstrates superior performance through experiments on anomaly detection benchmarks and ablation studies.

\section{Proposed framework}
\label{sec:Methodology}


The pipeline of our framework  is illustrated in Fig. \ref{fig:architecture}. It leverages a Bi-LSTM framework \cite{graves2005framewise} as the backbone, where  two enhanced ConvLSTM-AEs are used to  perform frame-to-frame prediction  in forward and backward  order. The outputs of both ConvLSTM-AEs  are then merged  to yield a more precise prediction. The anomalies are eventually identified based on prediction errors.

\begin{figure}[t]

\begin{minipage}[b]{1.0\linewidth}
  \centering
  \centerline{\includegraphics[width=8.5cm]{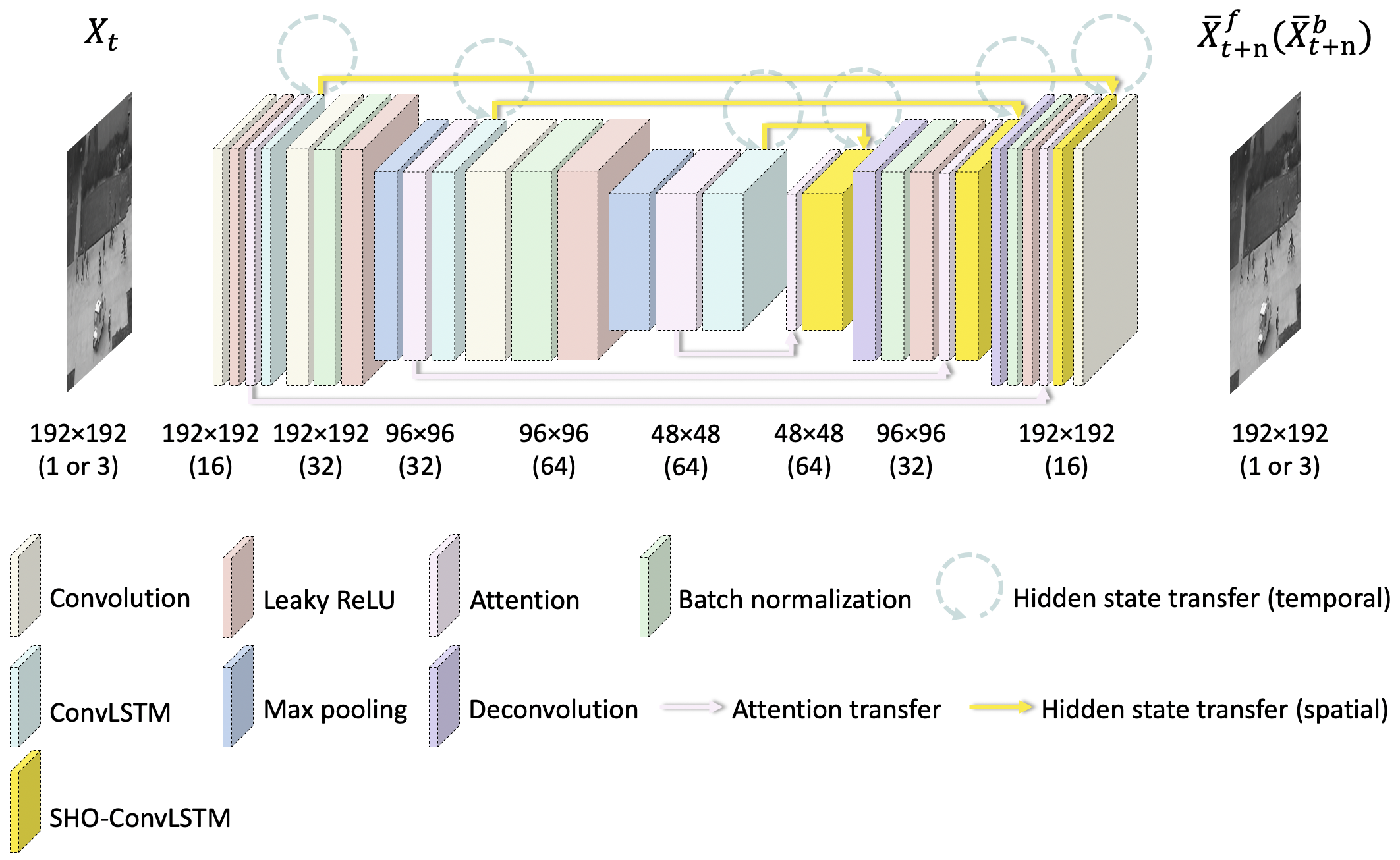}}
\end{minipage}

\caption{\small The architecture of the proposed ConvLSTM-AE. $X_{t}$ is a frame at timestep $t$. The output of the ConvLSTM-AE can be either $\bar{X}^f_{t+n}$ in forward pass or $\bar{X}^b_{t+n}$ in backward pass.}
\label{fig:ae}
\vspace{-2mm}
\end{figure}

As shown in Fig. \ref{fig:ae}, each ConvLSTM-AE consists primarily of Conv, Deconv and ConvLSTM layers that discover temporal dependencies on spatial data. They are stacked alternately to capture long-range features. Our spatial higher-order ConvLSTMs, hereinafter referred to as SHO-ConvLSTM, replace the original ConvLSTMs in the decoder to incorporate hidden states from the current encoder.  In addition, the  attention modules are implemented before ConvLSTM layers to accentuate the dynamics between frames  and overcome the drawbacks of the restricted receptive fields in ConvLSTMs.



\noindent\textbf{Bi-directionality.} Using a rigid temporal order in recurrent frameworks can potentially reduce the flexibility of temporal memory interaction \cite{schuster1997bidirectional}. To lift the restriction on  temporal order, we introduce a bi-directional mechanism derived from the original Bi-LSTM  \cite{graves2005framewise} in NLP. While the traditional Bi-LSTMs rely on past and future data, our mechanism focuses more on  the bi-directionality of past frames  for real-time use. As shown in Fig. \ref{fig:architecture}, the input sequence $\{X_{1},X_{2},...,X_{T}\}$ of length $T$  flows into two ConvLSTM-AEs, which conduct the forward and backward passes separately. The forward pass propagates the hidden states from timestep $1$ to $T$ and sequentially generates the forward predictions $\{\bar{X}^f_{1+n},\bar{X}^f_{2+n},...,\bar{X}^f_{T+n}\}$, where $n$ denotes the $n$-th frame after the input frame. Conversely, the backward pass conveys the hidden states from timestep $T$ to $1$ and generates the predictions in reverse order, $\{\bar{X}^b_{T+n},\bar{X}^b_{T+n-1},...,\bar{X}^b_{1+n}\}$. Both predictions  are then concatenated along the channel dimension and  mapped back to the image space to produce the final predictions $\{\bar{X}_{1+n},\bar{X}_{2+n},...,\bar{X}_{T+n}\}$. One can consider the process as implicit multi-task learning, where the framework is  trained to capture the temporal dimension in two complementary tasks, concurrently.

\begin{figure}[t]

\begin{minipage}[b]{1.0\linewidth}
  \centering
  \centerline{\includegraphics[width=8.5cm]{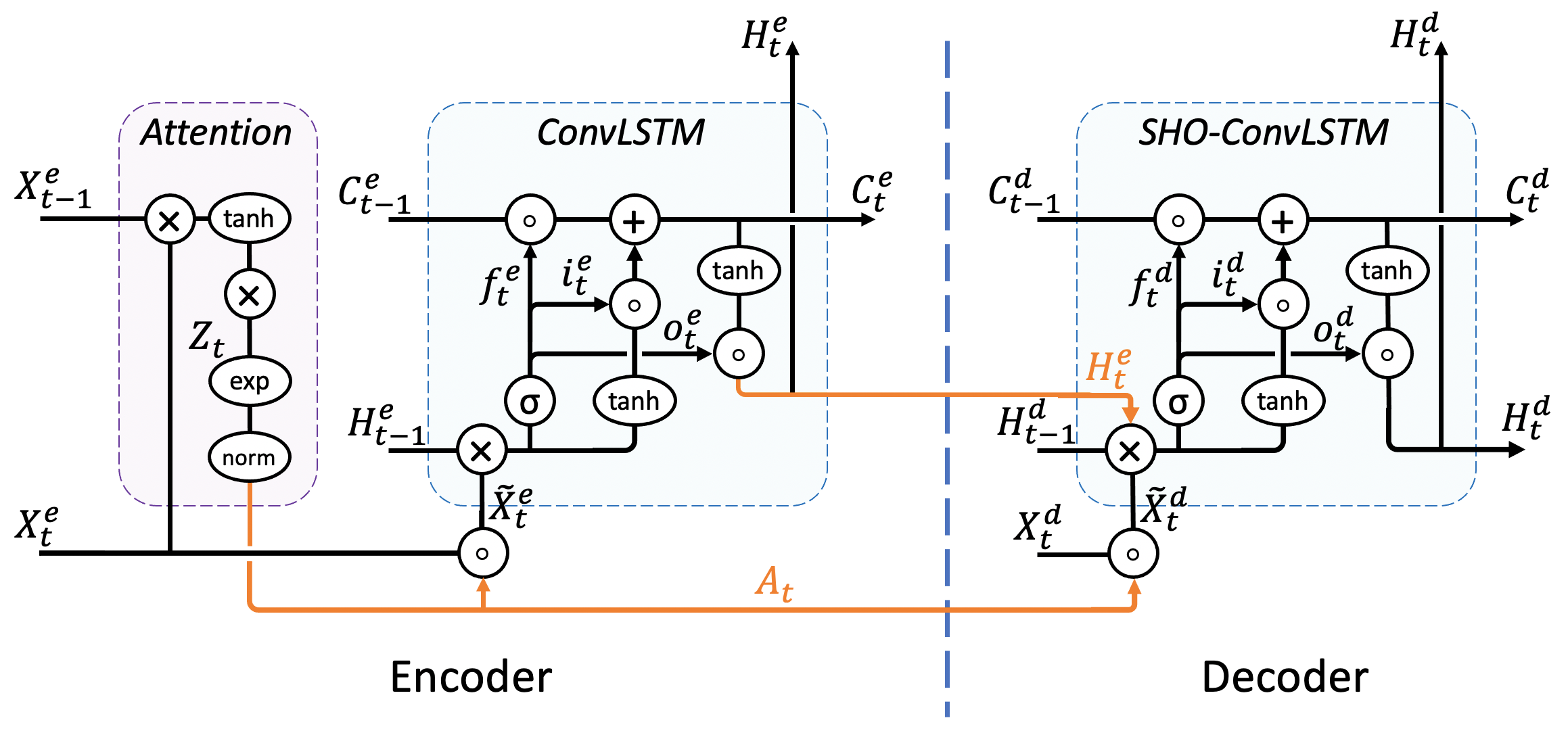}}
\end{minipage}

\caption{\small The spatial higher-order mechanism and  attention module. Superscripts $^e$ and $^d$ denote elements of the encoder and decoder, respectively; $H^{e}_t$ denotes the spatial memory flow between ConvLSTM and SHO-ConvLSTM;  $A_t$ denotes the attention mask.}
\label{fig:shoconvlstm}
\vspace{-2mm}
\end{figure}

\noindent\textbf{SHO-ConvLSTM.}   Due to the first-order Markovian nature, the traditional ConvLSTM only allows hidden states to be transferred over time but spatially independently, ignoring that spatial transitions can be equivalently essential to generate future frames \cite{wang2017predrnn}. Thus, we derive the SHO-ConvLSTM 
to enable spatial memory flows (see Fig. \ref{fig:shoconvlstm}), which is formulated as follows:
\begin{equation}
\small
\begin{aligned}
&\bar{C}^{d}_{t}=\tanh \left(W_{c} \otimes \left[\tilde{X}^{d}_{t}, H^{d}_{t-1}, \boldsymbol{H^{e}_{t}}\right]+b_{c}\right), \\
&i^{d}_{t}=\sigma\left(W_{i} \otimes \left[\tilde{X}^{d}_{t}, H^{d}_{t-1}, \boldsymbol{H^{e}_{t}}\right]+b_{i}\right), \\
&f^{d}_{t}=\sigma\left(W_{f} \otimes \left[\tilde{X}^{d}_{t}, H^{d}_{t-1}, \boldsymbol{H^{e}_{t}}\right]+b_{f}\right), \\
&o^{d}_{t}=\sigma\left(W_{o} \otimes \left[\tilde{X}^{d}_{t}, H^{d}_{t-1}, \boldsymbol{H^{e}_{t}}\right]+b_{o}\right), \\
&C^{d}_{t}=f^{d}_{t} \circ C^{d}_{t-1}+i^{d}_{t} \circ \bar{C}^{d}_{t}, \\
&H^{d}_{t}=o^{d}_{t} \circ \tanh \left(C^{d}_{t}\right),
\end{aligned}
\end{equation}
where $\tilde{X}^{d}_{t}$ is the input to the SHO-ConvLSTM at timestep $t$; $H^{d}_{t-1}$ is the hidden state from timestep $t-1$; $H^{e}_{t}$ is the hidden state from the corresponding ConvLSTM (of the same size) in the encoder at timestep $t$; $\bar{C}^{d}_{t}$, $i^{d}_{t}$, $f^{d}_{t}$ and $o^{d}_{t}$ denote the cell, input, forget and output states at timestep $t$, respectively; $\sigma(\cdot)$ denotes the Sigmoid function; $\circ$ and $\otimes$ denote the Hadamard product and convolution operations, respectively; and   $W$ and $b$ are trainable parameters of the SHO-ConvLSTM. The symbols in bold highlight the difference between a ConvLSTM and the SHO-ConvLSTM.

As shown in Fig. \ref{fig:shoconvlstm}, while the  ConvLSTM in the encoder still follows the standard ConvLSTM routine, the SHO-ConvLSTM in the decoder considers both the temporal dynamics $H^{d}_{t-1}$ and the spatial dynamics $H^{e}_{t}$. Hence, the decoder can efficiently retrieve information from both historical timesteps and the current encoder, as well as alleviate the gradient vanishing problem.

\noindent \textbf{Attention.}  Since ConvLSTMs alone cannot fulfill informative feature selection \cite{li2018videolstm}, we introduce an auxiliary attention module to estimate the importance weights of different elements in the feature maps (see Fig. \ref{fig:shoconvlstm}). The intuition behind the module is that the anomalies of interest  typically cause distinctive variations in  pixel values over time. To  focus on the variations, we compute the attention weights $Z_{t}$ by encoding both the current input $X^{e}_{t}$ and the previous input $X^{e}_{t-1}$:
\begin{equation}
\small
\begin{array}{c}
Z_{t}=W^{2}_{z} \otimes \tanh \left(W^{1}_{z}\otimes \left[ X^{e}_{t},  X^{e}_{t-1}\right]+b_{z}\right).
\end{array}
\end{equation}
Different from the schemes of convolving $X^{e}_{t}$ with the  hidden state $H^{e}_{t-1}$ \cite{li2018videolstm,zhang2018attention}, the proposed module does not rely on the ConvLSTM's output, and thus can achieve parallel computing. Meanwhile, it eases  feature fusion and extraction by taking the inputs with the identical context definition. The final attention mask $A_{t}$ is then generated for each feature map using \emph{min-max} normalization:
\begin{equation}
\small
A_{t}^{i j}=\frac{\exp \left(Z_{t}^{i j}\right)-\min _{i, j} \left(\exp \left(Z^{ij}_{t}\right)\right)}{\max_ {i, j} \left(\exp \left(Z^{ij}_{t}\right)\right)-\min _{i, j} \left(\exp \left(Z^{ij}_{t}\right)\right)},
\end{equation}
where $(i,j)$ represents the element's position in the map; the $\max_{i, j}(\cdot)$ and $\min_{i, j}(\cdot)$ operations, respectively, locate the maximum and minimum elements in $Z_{t}$. Compared to \emph{sum}, i.e., \emph{Softmax}, or \emph{max-only} normalization \cite{zhang2018attention}, \emph{min-max} allows for multiple maxima and minima, and enlarges the disparity between them. We hence expect the features that contribute more to the prediction to gain larger weights and vice versa. 

As shown in Fig. \ref{fig:shoconvlstm}, the attention mask $A_{t}$ is computed from the feature maps in the encoder and then imposed on both the ConvLSTM and the SHO-ConvLSTM using the Hadamard product, assuming that regions of interest only shift slightly in time and space.


\noindent\textbf{\textit{\textbf{SSIM +}} $\boldsymbol{ \ell_{1}}$ loss.}   The classic Manhattan ($\ell_{1}$) loss encourages sparsity in nonstructural  features and linearly mitigates the impact of outliers:
\begin{equation}\label{eq:l1}
\small
\mathcal{L}^{\ell_{1}}(P,\hat{P})=\frac{1}{|P|} \sum_{i, j \in P}\left|p_{ij}-\hat{p}_{ij}\right|,
\end{equation}
where $P$ and $\hat{P}$ denote the ground-truth  and its prediction,  respectively; $p_{ij}$ and $\hat{p}_{ij}$ are the pixel values of $P$ and $\hat{P}$ at position $(i,j)$, respectively; and $|P|$ is the number of pixels. However, the $\ell_{1}$ loss minimizes uncertainty by averaging all the probable solutions, i.e., by blurring the predictions. 
The structural similarity index measure \textit{(SSIM}) loss \cite{wang2004image}, on the other hand, evaluates perceivable differences in structure, luminance and contrast, and thus promotes visually convincing predictions:
\begin{equation}
\small
\mathcal{L}^{ssim}(P, \hat{P})=\frac{\left(2 \mu_{P} \mu_{\hat{P}}+c_{1}\right)\left(2 \sigma_{P \hat{P}}+c_{2}\right)}{\left(\mu_{P}^{2}+\mu_{\hat{P}}^{2}+c_{1}\right)\left(\sigma_{P}^{2}+\sigma_{\hat{P}}^{2}+c_{2}\right)},
\end{equation}
where $c_{1}$ and $c_{2}$ are the stabilizers; $\mu_{P}$ and $\mu_{\hat{P}}$ are the expectations of $P$ and $\hat{P}$, respectively; $\sigma^{2}_{P}$ and $\sigma^{2}_{\hat{P}}$ are the corresponding variances; and $\sigma_{P \hat{P}}$ is the covariance. Unfortunately, unlike the $\ell_{1}$ loss,  the \textit{SSIM} loss suffers from the  insensitivity to nonstructural distortions (e.g., shifting, scaling).

To benefit from both  losses, we  merge them to obtain the final objective function:
\begin{equation}
\small
\mathcal{L}^{mix}= \mathcal{L}^{ssim}+\lambda ( W_{\ell_{1}} \otimes \mathcal{L}^{\ell_{1}}+ b_{\ell_{1}}),
\end{equation}
where the weight $\lambda$ is pragmatically set to 1. Different from \cite{zhao2016loss}, an extra Gaussian filter parameterized by $W_{\ell_{1}}$ and $b_{\ell_{1}}$ is applied to the $\ell_{1}$ loss to ensure  coherence with the Gaussian-based \textit{SSIM} process.



\noindent\textbf{Anomaly inference.}  To estimate anomaly scores, 
we adopt the Mean Absolute Error (\textit{MAE}, i.e., $\ell_{1}$) metric as it partially conforms to
our objective function. The \textit{MAE} value is calculated for each frame $t$ using Eq. \ref{eq:l1}  and then normalized over each video to the range $[0,1]$ by using:
\begin{equation}
\small
S(t)=\frac{\mbox{\textit{MAE}}(t)-\min _{t} \left(\mbox{\textit{MAE}}(t)\right)}{\max _{t} \left(\mbox{\textit{MAE}}(t)\right)-\min _{t}\left(\mbox{\textit{MAE}}(t)\right)},
\end{equation}
where $\max_{t}(\cdot)$ and $\min_{t}(\cdot)$ determine the extreme \textit{MAE} values in each video. Large $S(t)$ values correspond to abnormal frames.

\section{EXPERIMENTAL RESULTS}
\label{sec:experimentalresults}


\noindent\textbf{Implementation details.} We evaluate  our framework on three public benchmarks: the UCSD Ped2 \cite{mahadevan2010anomaly}, the CUHK Avenue \cite{lu2013abnormal}, and the ShanghaiTech \cite{luo2017revisit}  datasets. All input frames are re-sized to $192\times192$ and normalized to the intensity range $[-1,1]$.  To verify the framework's robustness to different video sources, we preserve the original color format of each dataset: grayscale and RGB. The length $T$ of each  sample clip is fixed to $9$  since  anomalies usually occur in the short term. The receptive fields of the attention, Conv and Deconv layers are set to $3\times3$. The receptive fields of ConvLSTM and SHO-ConvLSTM  layers are set to $5\times5$ \cite{xingjian2015convolutional}. Leaky ReLU is used as the activation function. Batch normalization is inserted right after Conv and Deconv layers as the stabilizer. Adam, with an initial learning rate of  $5.0\times10^{-4}$, serves as the parameter optimizer.  We create a validation set from the original training set with a ratio $1:9$, and execute early stopping when the validation loss reaches an optimal value. To differentiate the prediction ability from the reconstruction ability, instead of simply predicting the next frame \cite{liu2018future,lai2020video}, our framework predicts the 7th frame after each input frame ($n=7$) and regards the 5th frame in the prediction sequence (i.e., the 3rd frame after the last input frame) as the expected result during testing.

\noindent \textbf{Metrics.} We utilize the frame-level Area Under Curve (AUC) of the Receiver Operating Characteristics (ROC) curve to quantitatively assess our framework against the SOTA. A higher AUC value suggests a higher probability that the framework will correctly detect abnormal frames.
 



\noindent\textbf{Comparison with the SOTA.} As shown in Table \ref{table:Peformances}, we compare our framework with several prediction-based SOTA methods.  Note that only unsupervised approaches are considered since supervised methods usually acquire knowledge from testing videos, which contradicts typical application scenarios. It is evident that the proposed framework outperforms most competitors on all three benchmarks.  Even on the most challenging ShanghaiTech dataset, it reports an overall score of 79.7\%, exceeding the  SOTA performance by 3.5\%.

\begin{table}
\small
\begin{center}
\caption{\small AUC (\%) of different models on three public benchmarks.}
\label{table:Peformances}
\setlength{\tabcolsep}{4pt}
\begin{tabular}{lccc}
\specialrule{.1em}{.05em}{.05em} 
Authors & UCSD Ped2 & CUHK Avenue & ShanghaiTech\\
\hline
 Luo et al. \cite{luo2017remembering}& 88.1 & 77.0 & - \\
 Luo et al. \cite{luo2017revisit} & 92.2 & 81.7 & 68.0\\
  Wang et al. \cite{wang2018abnormal} & 88.9 &\textbf{90.3} & - \\
  Liu et al. \cite{liu2018future}  & 95.4 & 85.1 & 72.8\\
   Lee et al. \cite{lee2019bman}  & \textbf{96.6} & 90.0 & \textbf{76.2}\\
   Song et al. \cite{song2019learning}       & 90.3 & 89.2 & 70.0\\
    Chen et al. \cite{chen2020anomaly}           & \textbf{96.6}& 87.8 & - \\
      Lai et al. \cite{lai2020video}  & 95.8 & 87.4 & - \\
\hline
Our framework & \textbf{98.3} & \textbf{90.7} & \textbf{79.7} \\
\specialrule{.1em}{.05em}{.05em} 
\end{tabular}
\end{center}
\vspace{-5mm}
\end{table}


\noindent \textbf{Ablation studies.} We study the impact of different components in our framework: bi-directionality, spatial higher-order mechanism, and attention module. To this end, we create several variants of the proposed framework by removing these components. The experiments are conducted on the CUHK Avenue dataset and assessed in terms of convergence (i.e., optimal validation loss) and AUC. As shown in Table \ref{table:eachcomponent}, each component  contributes individually to the framework’s performance for both metrics. Particularly, the bi-directionality and spatial higher-order mechanism significantly enhance the AUC score by 2.6\% ($\mathbb{A}$ vs.  $\mathbb{D}$) and 2.1\% ($\mathbb{B}$ vs.  $\mathbb{D}$),  respectively, proving the  effectiveness  of strengthening  memory exchange in ConvLSTMs.
 
\begin{figure}[t]

\begin{minipage}[b]{1.0\linewidth}
  \centering
  \centerline{\includegraphics[width=8.6cm]{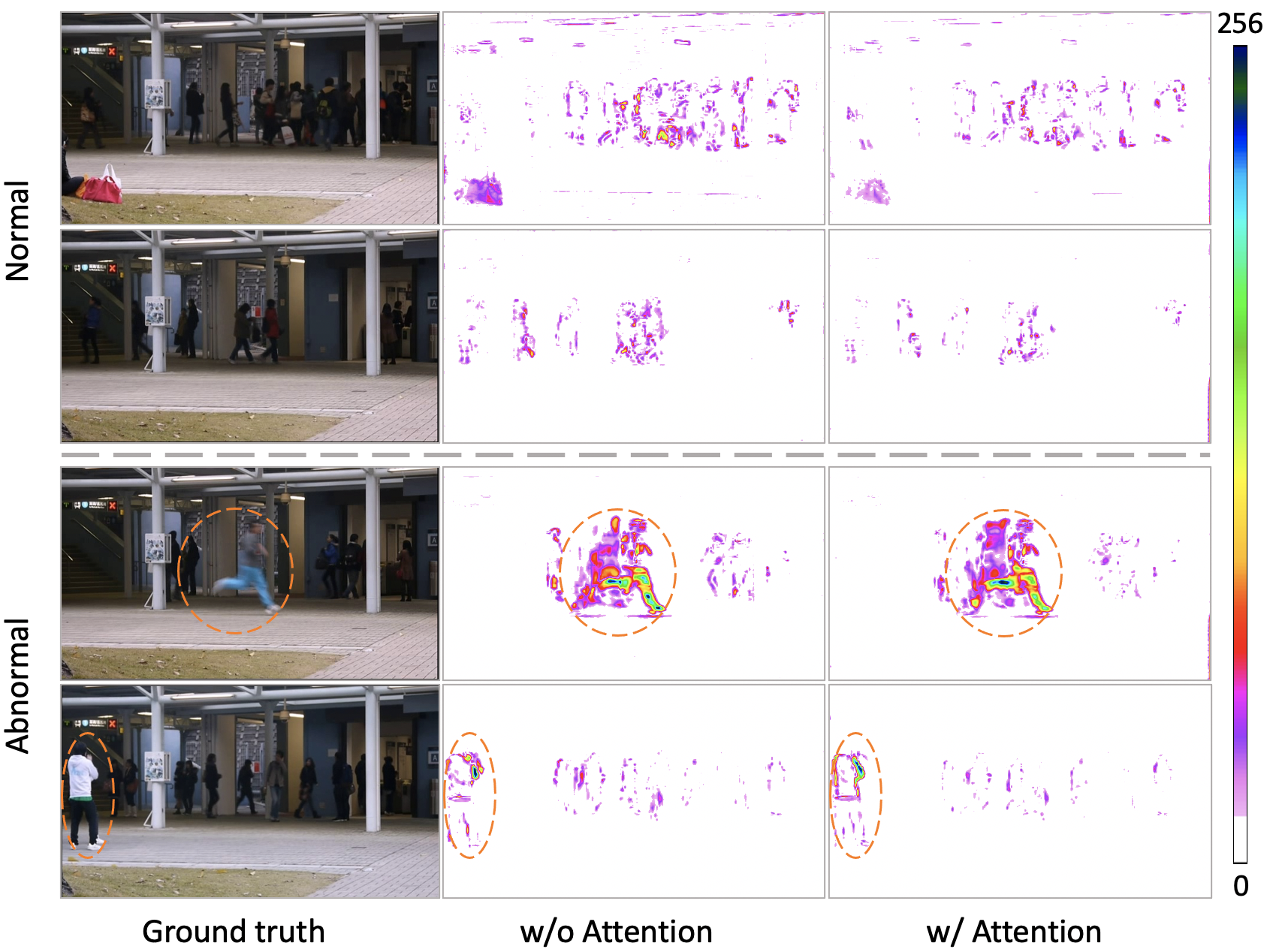}}
\end{minipage}

\caption{\small Examples of the attention module’s impact on normal and abnormal frames. The error maps in the second and third column are computed using $|\text{prediction} - \text{ground truth}|$. The dashed circles highlight the anomalies.}
\label{fig:attention}
\vspace{-2mm}
\end{figure}

 Fig. \ref{fig:attention} depicts qualitative examples of the attention module’s responses to different frames.  By analyzing  the error maps between prediction and ground truth, we observe that our attention mechanism can diminish the prediction error around normal moving items, while preserving it in abnormal areas. This confirms that the attention module can help the framework to focus on more predictive features and correctly predict normal patterns.

\begin{table}
\small
\begin{center}
\caption{\small Evaluation of different components of our framework on the CUHK Avenue dataset.}
\label{table:eachcomponent}
\setlength{\tabcolsep}{3.5pt}
\begin{tabular}{>{\centering}p{0.05\textwidth}>{\centering}p{0.05\textwidth}>{\centering}p{0.05\textwidth}>{\centering}p{0.05\textwidth}>{\centering}p{0.13\textwidth}p{0.07\textwidth}}
\specialrule{.1em}{.05em}{.05em} 
\multirow{2}{*}{\small Index}&\multicolumn{3}{c}{\small Framework design} & \multirow{2}{*}{\small Val loss ($\times10^{-3}$)} &\multirow{2}{*}{\small AUC (\%)} \\\cline{2-4}

 &Bi & SHO & Att &  & \\
\specialrule{.1em}{.05em}{.05em} 
 $\mathbb{A}$   &\ding{55}&\ding{51} & \ding{51}&17.58 &\hfil 88.1\\
   $\mathbb{B}$ &\ding{51}&\ding{55} &\ding{51} &10.49 & \hfil 88.6\\
   $\mathbb{C}$  &\ding{51}&\ding{51} &\ding{55} &4.62 &\hfil 89.8 \\
   $\mathbb{D}$   &\ding{51}&\ding{51} &\ding{51} & 4.54 &\hfil 90.7  \\

\specialrule{.1em}{.05em}{.05em} 
\multicolumn{5}{l}{\tiny Bi = Bi-directionality. SHO = Spatial higher-order. Att = Attention. Val loss = Validation loss. }
\end{tabular}
\end{center}
\vspace{-2mm}
\end{table}

\section{CONCLUSION}
\label{sec:conclusion}

We presented a spatio-temporal memory-enhanced  ConvLSTM-AE predictor to detect anomalies as the patterns that deviate from expectations. The framework's bi-directionality helps to effectively exploit the temporal dimension through the  backward prediction task. Moreover, its spatial higher-order ConvLSTMs can retrieve hidden states from the current encoder to boost spatial information exchange. Finally, its attention module helps to locate the features that are more informative for frame generation by comparing neighboring frames. Our ablation studies and experiments on public benchmarks confirm the effectiveness of each component and validate the strong performance of our framework.

\clearpage

\bibliographystyle{IEEEbib}
\bibliography{strings,ICASSP_SGD}

\end{document}